\documentclass{article}
\usepackage{ijcai24}

\usepackage{times}
\usepackage{soul}
\usepackage{url}
\usepackage[hidelinks]{hyperref}
\usepackage[utf8]{inputenc}
\usepackage[small]{caption}
\usepackage{graphicx}
\usepackage{amsmath}
\usepackage{amsthm}
\usepackage{booktabs}
\usepackage{algorithm}
\usepackage[switch]{lineno}

\usepackage{comment}
\usepackage{amsfonts}
\usepackage{subfigure}
\usepackage{kotex}
\usepackage{algpseudocode}

\title{Improving Diffusion-Based Generative Models\\via Approximated Optimal Transport}

\author{
    Author 4934
    \affiliations
    IJCAI2024 Main Track
    \emails
}
\iftrue
\author{
 Daegyu Kim$^1$
\and
Jooyoung Choi$^1$\and
Chaehun Shin$^1$\and
Uiwon Hwang$^2$\thanks{Corresponding Authors} \And
Sungroh Yoon$^{1*}$  \\
\affiliations
$^1$Data Science and AI Laboratory, ECE and Interdisciplinary Program in AI, Seoul National University\\
$^2$Division of Digital Healthcare, Yonsei University\\
\emails
\{large-scale, jy\_choi, chaehuny, sryoon\}@snu.ac.kr, 
uiwon.hwang@yonsei.ac.kr
}
\fi

\begin{document}

\maketitle

\begin{abstract}

We introduce the Approximated Optimal Transport (AOT) technique, a novel training scheme for diffusion-based generative models. Our approach aims to approximate and integrate optimal transport into the training process, significantly enhancing the ability of diffusion models to estimate the denoiser outputs accurately. This improvement leads to ODE trajectories of diffusion models with lower curvature and reduced truncation errors during sampling. We achieve superior image quality and reduced sampling steps by employing AOT in training. Specifically, we achieve FID scores of 1.88 with just 27 NFEs and 1.73 with 29 NFEs in unconditional and conditional generations, respectively. Furthermore, when applying AOT to train the discriminator for guidance, we establish new state-of-the-art FID scores of 1.68 and 1.58 for unconditional and conditional generations, respectively, each with 29 NFEs. This outcome demonstrates the effectiveness of AOT in enhancing the performance of diffusion models.

\end{abstract}

\section{Introduction}
Diffusion models~\cite{ho2020denoising,song2020denoising} refer to a group of generative models synthesizing images by progressive denoising, starting with Gaussian noise. Recent advancements in diffusion models demonstrate their ability to produce high-quality images, as seen in various studies~\cite{rombach2022high,ramesh2022hierarchical,saharia2022photorealistic}. 
    A key aspect of diffusion models is learning the score function~\cite{song2020denoising} to synthesize images.
    The score function represents the log-likelihood gradient of data distribution and is acknowledged as a pivotal operation within diffusion models by Song \textit{et al.}~\shortcite{song2021scorebased}. 
These models generate images by solving stochastic differential equations (SDE) or ordinary differential equations (ODE), using the model outputs as gradients for numerical integration, a crucial step in the image synthesis process.

    \textbf{EDM}~\cite{karras2022elucidating}, recognized for its effective image synthesis capabilities, observes that models with straightened ODE trajectories can synthesize high-quality images in fewer iterations. This observation significantly improves diffusion models through techniques to reduce the curvature of ODE and truncation error. As a result, these models achieve state-of-the-art performance, synthesizing high-quality images in fewer steps than previous research in this field.
    Additionally, several studies~\cite{kim2022refining,song2023consistency,zhang2023contrastive,kim2023consistency} have demonstrated high performance by adapting their approaches to well-performing pre-trained EDM models.

\begin{figure}
    \centering
    \includegraphics[width=0.98\columnwidth]{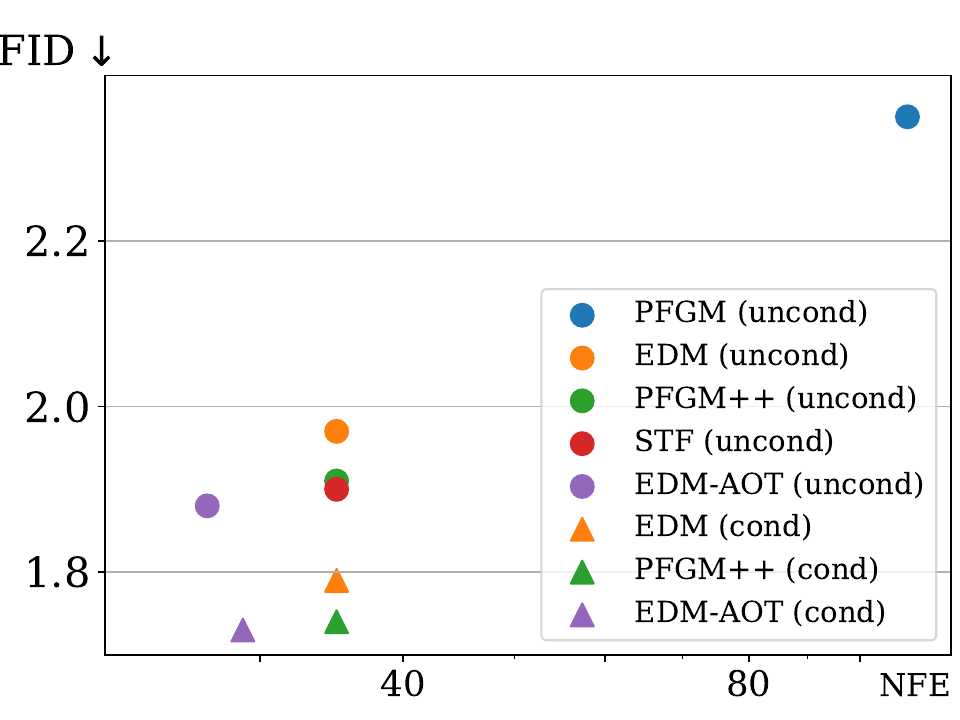}
    \vspace{-1em}
    \caption{
    Comparison of FID scores and the corresponding number of function evaluations (NFEs) for CIFAR-10 image unconditional and conditional generations in baseline studies and EDM-AOT. This graph demonstrates the superior performance of our approach in terms of image quality and reduced NFE compared to the baseline.}
    \vspace{-1em}
    \label{fig_main}
\end{figure}

Despite advancements, diffusion models still exhibit intervals of high curvature in their ODEs, likely due to the increased information entropy in the training process. 
While Flow Matching (a family of generative models different from diffusion models) and related studies~\cite{lipman2022flow,tong2023conditional,pooladian2023multisample} have proposed methods to address curved ODE trajectories using optimal transport~\cite{villani2003topics}, applying these approaches to diffusion models presents a computational efficiency issue due to the structure of diffusion models.

To counter this, we developed a training technique incorporating Approximated Optimal Transport (AOT), specifically targeting these curvature issues in diffusion models.
We demonstrate the feasibility of replacing the computation of the optimal transport with matching pairs of images and noise.
We approximate the computation of optimal transport as an assignment problem~\cite{kuhn1955hungarian}.
We employ this AOT technique to select the specific noise to pair with the dataset images during the training process.
This alteration reduces the information entropy of the training target for diffusion models under high noise levels, resulting in straightened trajectories of the ODE.
Diffusion models trained with our AOT technique exhibit reduced sampling steps while maintaining high-quality performance.

Diffusion models trained with our AOT technique exhibit outstanding performance in image generation on the CIFAR-10 dataset~\cite{krizhevsky2009learning}, delivering high-quality results with a low number of function evaluation (NFE), as shown in Figure~\ref{fig_main}.
Specifically, we achieve Fréchet Inception Distance (FID) scores~\cite{heusel2017gans} of 1.88 with 27 NFEs in unconditional generation and 1.73 with 29 NFEs in conditional generation.
Also, models trained with AOT can apply advanced sampling techniques to improve the diffusion models, using Discriminator Guidance (DG)~\cite{kim2022refining}. 
We achieved state-of-the-art FID scores of 1.68 in unconditional generation and 1.58 in conditional generation by applying DG, both performed with 29 NFEs, with our minor adaptation during discriminator training.

\begin{figure}[t]
\centering
\includegraphics[width=0.98\columnwidth]{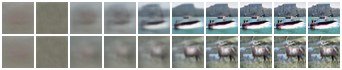}
\vspace{-.1em}
\caption{The images are generated using the unconditional EDM model with a single Euler step at each noisy image during the sampling process, proceeding from left to right. The images exhibit consistency, especially noticeable at low noise levels.}
\vspace{-1em}
\label{edm_ode_curv}
\end{figure}

\section{Preliminaries}

\subsection{Diffusion and Score Models}
Diffusion models~\cite{ho2020denoising} generate images through a reverse diffusion process, estimating and then reducing noise from a noisy image. This iterative denoising process enables these models to synthesize images from pure noise.
Song and Ermon~\shortcite{song2019generative} introduced score models which learn the log-likelihood gradient to synthesize images.
Subsequently, Score-based Generative Models (SGM)~\cite{song2021scorebased} integrate score models into the diffusion process for image generation.
They establish that the noise handled by diffusion models corresponds to the score.
These models use this score and numerical integration to solve stochastic differential equations (SDE) or ordinary differential equations (ODE) for image synthesis.
The processes of SDE, both forward and backward, can be described as follows:
\begin{gather}
d\mathbf{x} = \mathbf{f}(\mathbf{x},t) dt + g(t) d\mathbf{w}, \\
d\mathbf{x} = \left(\mathbf{f}(\mathbf{x},t) - g(t)^2 \mathbf{s}_\theta (\mathbf{x},t)\right) dt  + g(t)         d\Bar{\mathbf{w}},
\end{gather}
where $\mathbf{f}(\mathbf{x},t)$ and $g(t)$ serve as coefficients, and $\mathbf{w}$ and $\Bar{\mathbf{w}}$ correspond to the Wiener processes associated with the forward and backward processes, respectively.
A score-based model, denoted as $\mathbf{s}_\theta$, learns the gradient of a data distribution $\nabla_\mathbf{x} \log p_t(\mathbf{x})$.

Song\textit{ et al.}~\shortcite{song2021scorebased} proposed Probability Flow ODE (PF-ODE), which utilizes the same score models as those used in the SDE process for implicit image generation.
The equation governing the backward process of the ODE is as follows:
\begin{gather}
d\mathbf{x} = \left(\mathbf{f}(\mathbf{x},t) - \frac{1} {2} g(t)^2 \mathbf{s}_\theta (\mathbf{x},t)\right) dt.
\end{gather}
\subsection{EDM}
\label{2-2}
\begin{figure}[t]
\centering
\subfigure[]{\includegraphics[width=0.325\columnwidth]{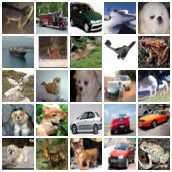}}
\subfigure[]{\includegraphics[width=0.325\columnwidth]{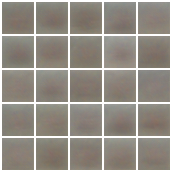}}
\subfigure[]{\includegraphics[width=0.325\columnwidth]{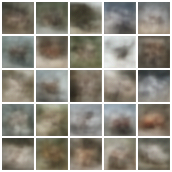}}
\vspace{-1em}
\caption{Synthesized CIFAR-10 images in an unconditional generation. (a) Denoised images $\mathbf{x}_0$ with EDM and its 35 NFEs sampler. (b) Denoised images $\mathbf{x}_0$ using the EDM model and single-step Euler method. (c) Denoised images $\mathbf{x}_0$ using the EDM-AOT model and single-step Euler method. The denoised images synthesized by EDM-AOT exhibit greater diversity than those synthesized by EDM.
}
\vspace{-1em}
\label{1step_sample}
\end{figure}
\begin{algorithm*}[t]
\caption{Training model $\theta$ with Approximated Optimal Transport.}
\label{alg}
\begin{algorithmic}[1]
\State \textbf{Initialize} model parameters $\theta$.
\State \textbf{Prepare} $p(\mathbf{y})$, $p(\sigma)$, $c$, and $N$. \Comment{ $p(\mathbf{y})$: dataset distribution, $p(\sigma)$: noise deviation distribution} 
\Repeat
\Comment{$c$: cost function, $N$: number of pairs}
    \State \textbf{Sample} $\{\mathbf{\mathbf{y}}_i\}_{i=1}^N \sim p(\mathbf{\mathbf{y}})$ 
    \State \textbf{Sample} $\{\epsilon_i\}_{i=1}^N \sim \mathcal{N}(0, I)$
    \State $\mathbf{y}' \leftarrow \text{Augmentation}(\mathbf{y})$ \Comment{Apply the augmentation. $\mathbf{y}'$ represents the augmented image.}
    \State Computing AOT through Hungarian algorithm: \Comment{Compute AOT between augmented images and random noise.}
    \State $\quad$ \textbf{Find} $\pi' = \arg\min_\pi \sum_{i=1}^N c (\epsilon^\pi_i, \mathbf{y}'_i) $ 
     \Comment{Find optimal indexes $\pi$, $\epsilon^\pi$ denotes the permutation following $\pi$.}
    \State $\quad $  $ \epsilon' \leftarrow \epsilon^\pi$ \Comment{Select noise $\epsilon'$ following the Hungarian algorithm as in Equation~\ref{equ9}.}
    \While{ until using all pairs $ (\mathbf{y}',\epsilon')$ once } \Comment{$B$ : number of mini-batch size per GPU}
    \State \textbf{Sample} $ \{(\bar{\mathbf{y}}_i ,\bar{\epsilon}_i )\}_{i=1}^B \sim (\mathbf{y}', \epsilon') $ 
    \Comment{Sample mini-batch from augmented pairs.}
    \State \textbf{Sample} $\{\sigma_i\}_{i=1}^B \sim p(\sigma)$ \Comment{$p(\sigma)$ conforms to the exponential value of a Gaussian distribution.}
    \State Compute loss: \Comment{$D_\theta$ estimates images from noisy input images and noise deviation.}
    \State $\quad L \leftarrow \frac{1}{B} \sum_{i=1}^B \lambda(\sigma_i)\| D_\theta (\bar{\mathbf{y}}_i + \sigma_i\bar{\epsilon}_i;\sigma_i) - \bar{\mathbf{y}}_i \|_2^2$ 
    \State Update parameters $\theta$ :
    \State $\quad \theta \leftarrow \theta - \eta \nabla L$ 
\EndWhile
\Until converged
\State \textbf{return} trained model $\theta$
\end{algorithmic}
\end{algorithm*}
\textbf{EDM}~\cite{karras2022elucidating} brings several enhancements to diffusion models, enabling the synthesis of high-quality images with fewer sampling steps than earlier models. 
This advancement forms the foundation of our experimental approach, leveraging EDM's image generation efficiencies.

\subsubsection{Model Training}

EDM's training aligns with conventional diffusion models, optimizing the following objective function:
\begin{gather}
\mathbb{E}_{p(\mathbf{y}),p(\sigma)} \lambda(\sigma)\|D_\theta(\mathbf{y} + \sigma \epsilon ; \sigma) - \mathbf{y}\|^2_2.
\label{equ:training}
\end{gather}
Here, $D_\theta$ represents the diffusion model, $p(\mathbf{y})$ the data distribution, $p(\sigma)$ the sampling distribution of $\sigma$, $\lambda(\sigma)$ the loss weight, and $\epsilon$ is random noise.
The choices made for these distributions and functions significantly enhance the overall performance of diffusion models.
\subsubsection{ODE Curvature Scheduling}

EDM identifies a correlation between lower ODE trajectory curvature and requirements of the sampling step requirements.
As trajectory curvature decreases, the consistency in the direction of the tangent increases, thereby reducing the truncation error when a high stride is applied.
It proposes noise schedules that leverage this straightforward ODE, as depicted in Figure~\ref{edm_ode_curv}.
EDM selects its noise schedule and sampling method to minimize curvature in the ODE trajectory.

\subsubsection{Efficient Sampling}

EDM introduces a superior sampling method utilizing Heun's method~\cite{ascher1998computer} and a novel time step selection strategy.
This approach reduces truncation errors and allows fewer sampling steps in high-quality image generation. Time steps are selected based on the following equation:
\begin{gather}
\label{equ-timesteps}
t_i = \left( {\sigma_\mathrm{max} }  ^\frac{1}{\rho} + \frac{i}{n-1}({\sigma_\mathrm{min}}^\frac{1}{\rho} - {\sigma_\mathrm{max}   }^\frac{1}{\rho} )\right)^\rho.
\end{gather}
Here, $n$ represents the number of steps, and $t_i$ signifies the selected time step value in the $i$-th step.
In this equation, $\rho$ adjusts the size of the strides.
A higher $\rho$ implies using wide strides near $\sigma_\mathrm{max}$ and narrow strides near $\sigma_\mathrm{min}$ to optimize generation performance; the EDM has used $\rho$ as 7 in their settings.
Given that Heun's method is a second-order method, the NFE during sampling with EDM is 2$n$-1.

\subsection{Optimal Transport for ODE-Based Models}
\label{2-3}

Flow Matching~\cite{lipman2022flow} is a recent family of ODE-based generative models.
These models synthesize images by integrating vectors of a trained vector field, similar to diffusion models.
Flow Matching has the advantage that these models are simulation-free.
Given that the vector field is the training target, these models can select the target they intend to simulate.

Tong\textit{ et al}.~\shortcite{tong2023conditional} introduced Optimal Transport Conditional Flow Matching (OT-CFM), which employs optimal transport~\cite{villani2003topics} to enhance performance by straightening the vector field trajectories. 
This methodology is similar to the EDM approach, which aims to reduce the curvature of ODE trajectories. 
Several studies~\cite{pooladian2023multisample,liu2023flow} have conducted investigations into straightening trajectories.

The optimal transport of two distributions can be determined using the 2-Wasserstein distance~\cite{villani2009optimal,arjovsky2017wasserstein}, represented by the following equation:
\begin{gather}
\label{OT_cost}
W(q_0,q_1)^2_2 = \underset{\pi \in \Pi}{\mathrm{inf}} \mathbb{E}_{(x,y)\sim \pi} c(x,y),
\end{gather}
where $\Pi$ denotes the joint probability matrix of two probability distributions, $q_0$ and $q_1$.
$c$ refers to the cost function quantifying the distance between two inputs.

Given that $u_t$ represents the vector field of Flow Matching and the cost function $c$ denotes Euclidean distance, the dynamic formulation of the 2-Wasserstein distance is expressed as follows:
\begin{gather}
\label{2-wass}
W(q_0,q_1)^2_2 = \underset{p_t, u_t}{\mathrm{inf}} \int_{\mathbb{R}^d} \int^1_0 p_t(\mathbf{x})\|u_t(\mathbf{x})\|^2 dt d\mathbf{x}, 
\end{gather}
where $p_t$ denotes the distribution of $\mathbf{x}$ at specific time $t$, within the ODE process.

This dynamic formulation (Equation~\ref{2-wass}) illustrates why applying optimal transport techniques to diffusion models is challenging.
As mentioned by Tong\textit{ et al.}~\shortcite{tong2023conditional}, computing the optimal transport probability requires integrating across all time steps.
This aspect is inconsequential in simulation-free models, such as Flow Matching. 
However, in the case of diffusion-based generative models, evaluating models at all time steps is necessary for Wasserstein distance computation.
This aspect results in computational inefficiency when employing optimal transport with diffusion models.

\subsection{Hungarian Algorithm}
\label{2-4}

\subsubsection{Assignment Problem}

The assignment problem~\cite{kuhn1955hungarian} entails finding the optimal matching between agents and tasks, aiming to minimize the associated costs.
In this context, each agent bears a unique cost associated with performing each task, and each agent is allocated a single task without repetition.
The assignment problem can be interpreted in the context of optimal transport with finite distribution settings.

\subsubsection{Hungarian Algorithm}
Hungarian algorithm~\cite{kuhn1955hungarian} is an algorithm to search the solution of the assignment problem with time complexity $O(n^3)$, where $n$ denotes the number of agents or tasks.
This algorithm leverages the property that the optimal solution remains unchanged when the exact value is added or subtracted from every cost associated with the same agent or task.
The algorithm finds the optimal solution through iterations of the technique above, ensuring that all pairs' total cost equals 0.
Our approaches utilize the algorithm to search for the optimal transport of finite distributions between sampled images and noise.

\begin{table}[t]
\centering
\begin{tabular}{lcccc}
\toprule
    Models & NFE $\downarrow$ & FID $\downarrow$  \\ 
    \midrule
    DDPM~\cite{ho2020denoising}&1000&3.17\\
    SGM~\cite{song2021scorebased}&2000&2.20\\
    PFGM~\cite{xu2022poisson}&110&2.35\\
    LSGM~\cite{vahdat2021score}& 138&2.10\\
    EDM~\cite{karras2022elucidating} & 35 & 1.97 \\
    PFGM++~\cite{xu2023pfgm++}& 35 & 1.91 \\
    STF~\cite{xu2023stable} & 35 & 1.90 \\ 
    
    \midrule
    EDM-AOT ($\rho=7$, 18 steps) & 35 & 1.95\\
    EDM-AOT ($\rho=90$, 14 steps) & \textbf{27} & \textbf{1.88} \\
    \bottomrule
\end{tabular}
\caption{Comparison with the baseline results of other diffusion-based generative models for unconditional CIFAR-10 generation. EDM-AOT employing appropriate sampling hyper-parameters achieves the lowest FID score with fewer NFEs.}
\vspace{-1em}
\label{baseline}
\end{table}

\section{Motivation}
 As highlighted by Karras\textit{ et al.}~\shortcite{karras2022elucidating}, the curvature of the ODE trajectory plays a crucial role in the performance of diffusion models. 
 Figure~\ref{edm_ode_curv} presents an analysis of the ODE trajectory in EDM.
 The ODE trajectory exhibits low curvature at low noise levels, resulting in images estimated using single Euler steps that closely match the target with minimal deviations. This alignment suggests a more downward curvature in these regions of the ODE trajectory.

Conversely, in segments of high noise levels, as shown in Figure~\ref{1step_sample}-(b), the model's estimates show too blurry images, suggesting elevated information entropy.
In these high noise regions, the initial estimates are distant from the sampled images, necessitating the model's iterative refinements. This phenomenon indicates an increased curvature in the ODE process, as depicted in Figure~\ref{edm_ode_curv}.

As the models tend to estimate the posterior mean given noisy data $\mathbb{E}[\mathbf{x}_0|\mathbf{x}_t,t]$, the model learns to average over all possible random noise perturbations under the high noise scale, resulting in increased information entropy during the training process. 
To effectively address this, we propose a training technique detailed in Section~\ref{sec4}, which enables models to learn the lower curvature ODE trajectory. 
Drawing inspiration from the concepts of Flow Matching (Section~\ref{2-3}), our goal is to reduce the information entropy in the training process, especially under high noise levels.

\section{Approximated Optimal Transport Training}
\label{sec4}

This section presents a novel training technique of diffusion models dubbed Approximated Optimal Transport. 
We summarize the detailed training process in Algorithm~\ref{alg}.

\subsection{Training Process using AOT}

In the training process of standard diffusion models, the synthesis of noisy images is typically achieved by using random noise independent of the target image. These models are conventionally trained following Equation~\ref{equ:training}, where the target images, denoted as $\mathbf{y}$, and the noise $\epsilon$ are sampled independently. This method forms the foundation of many existing diffusion models and is a well-established approach in the field.

We propose modifying this standard process with our ``Approximated Optimal Transport" (AOT) technique. Instead of relying on randomly sampled noise, AOT employs selectively chosen Gaussian noise $\epsilon'$ that bears a closer relationship to the target image. This alteration in noise selection enhances the connection between the target images and the noise, leading to a more effective learning process, thereby reducing the information entropy. By replacing the conventional random noise sampling with this tailored approach, AOT marks a notable shift from traditional practices in diffusion models. 
We will discuss The detailed implementation and implications of AOT in Section~\ref{4-2}.

\begin{table}[t]
\centering
\begin{tabular}{lcccc}
\toprule
    Models & NFE $\downarrow$ & FID $\downarrow$  \\ 
    \midrule
    EDM~\cite{karras2022elucidating} & 35 & 1.79 \\
    PFGM++~\cite{xu2023pfgm++}& 35 & 1.74 \\
    \midrule
    EDM-AOT ($\rho=7$, 18 steps) & 35 & 1.79\\
    EDM-AOT ($\rho=72$, 15 steps) & \textbf{29} & \textbf{1.73} \\
    \bottomrule
\end{tabular}
\caption{Comparison with the baseline results of other diffusion-based generative models for conditional CIFAR-10 generation. EDM-AOT employing appropriate sampling hyper-parameters achieves the lowest FID score with fewer NFEs.}
\vspace{-1em}
\label{baseline-cond}
\end{table}

\begin{figure*}[t]
\centering

\subfigure[]{\includegraphics[width=0.33\textwidth]{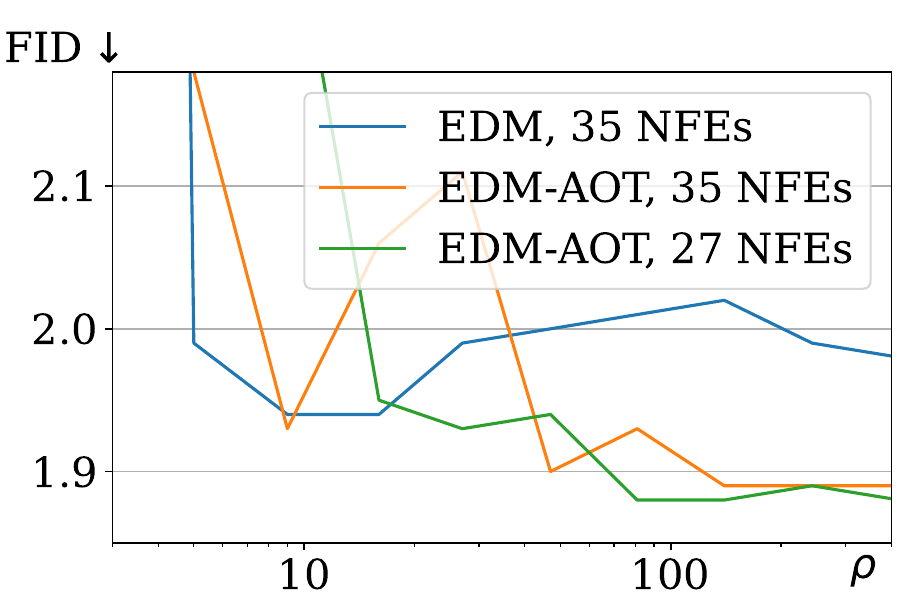}}
\subfigure[]{\includegraphics[width=0.33\textwidth]{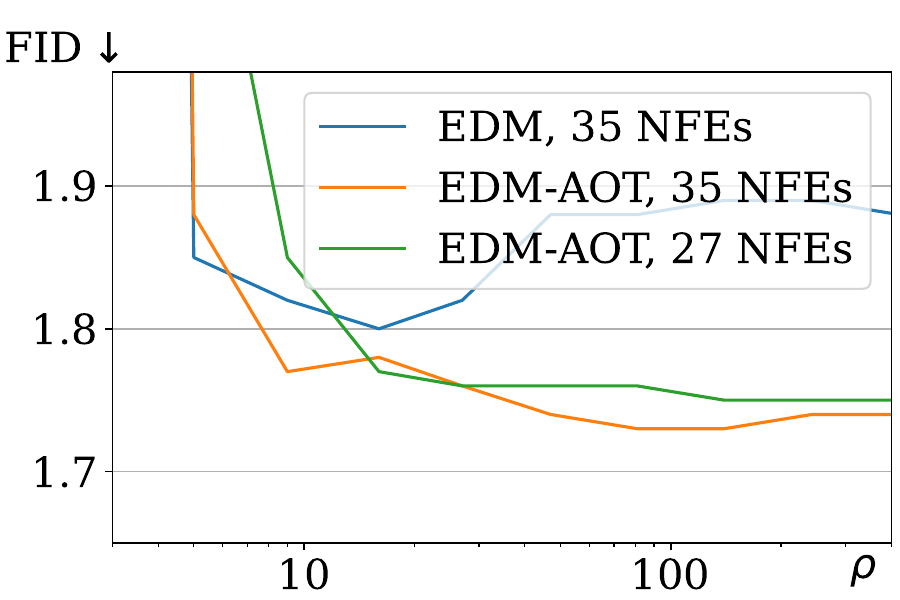}}
\subfigure[]{\includegraphics[width=0.33\textwidth]{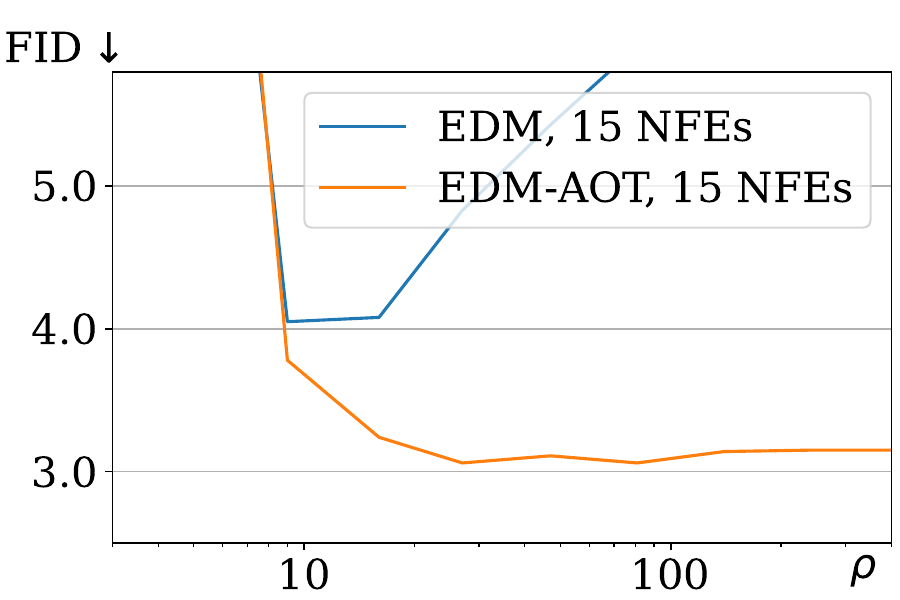}}
\vspace{-1em}
\caption{FID scores of CIFAR-10 images across varying $\rho$ values and model configurations. The models employing AOT exhibited consistent performance even as $\rho$ increased. (a) FID scores in unconditional generation. (b) FID scores in conditional generation. (c) FID scores in unconditional generation, with 8 steps (15 NFEs).
}
\vspace{-1em}
\label{rho_ablation}
\end{figure*}

\subsection{AOT Implementations}
\label{4-2}
\subsubsection{Computation Process of AOT}

We aim to determine the specific $\epsilon'$ corresponding to the optimal transport in the diffusion model training process.
To determine the optimal transport route, we aim to calculate the 2-Wasserstein distance between two distributions as described in Equation~\ref{2-wass}. 
Given that the distribution of the real image dataset is not tractable, computing the optimal transport solution between the target image and noise becomes challenging.

We approximate the optimal transport between the two distributions at the batch level to address this challenge.
Rather than determining the optimal transport for two distributions, we identify pairs with the smallest sum of cost functions in each iteration. 
In each iteration, we draw a sample of $N$ target images $\mathbf{y}$ from the dataset, paired with $N$ random noise, where $N$ denotes the number of pairs to search the AOT.
We also sample $N$ random noises in each iteration and identify the transport function using only these two sets of samples.

In the search for the optimal transport, the objective is to find a transport function that minimizes the sum of the cost function for all elements (Equation~\ref{OT_cost}).
In our case, it is necessary to determine this transport for only $N$ pairs of elements. 
We are in search of the indexing function $\pi$ that minimizes the costs, and the following expression encapsulates this pursuit:
\begin{gather}
    \arg\min\limits_\pi \sum\limits_{i=1}^N c(\epsilon_{\pi(i)},\mathbf{y}_i).
    \label{equ9}
\end{gather}
Here, $\pi(i)$ denotes a function that assigns each $i$ to a unique index, while $\epsilon_i,$ and $ \mathbf{y}_i$ denote the $i$-th selected noise and target images.
To address this objective, we frame it as the assignment problem and solve it with the Hungarian algorithm~\cite{kuhn1955hungarian}, as mentioned in section~\ref{2-4}.

First, we construct a distance matrix denoted as $W$, in which $W_{ij}$ represents the cost value between the $i$-th target image and the $j$-th noise.
In our process, we use L2 distance as the cost function.
Subsequently, we determine the index of the optimal pair matching using the Hungarian algorithm.
We then utilize these determined pairs to synthesize noisy images for training the diffusion models.

In each iteration, we choose selected noise $\epsilon'$ from the sampled noise to ensure that the $\mathbf{x}_\sigma$ distribution corresponds to the original EDM. 
We imply that selecting noise not from the sampled noise could compromise the alignment of the distribution of the chosen noise with the Gaussian distribution.

\subsubsection{AOT in Conditional Generation}
When training models with labeled data, such as the CIFAR-10 dataset~\cite{krizhevsky2009learning} with class-conditional generation, we determine selected noise $\epsilon'$ to each set of images that share the same labels.
We divide $N$ target images into $C$ sets sharing the same labels, where $C$ denotes the number of classes of labels.
We then sample $N$ noise and divide them into $C$ sets, ensuring that each set's numbers of target images and noise are the same.
We construct $C$ distance matrices similar to unconditional generations and apply the Hungarian algorithm to each matrix.

We determine the pairs to each label to remove the dependency between the distributions of labels and noise.
If correlations exist, the processes for training and sampling may vary.
We randomly sampled the label and Gaussian noise to synthesize images but could not account for their correlations.
We eliminate these correlations through additional adaptation to ensure consistency between training objectives and initial sampling conditions.
We provide the detailed algorithm in the supplementary.

\subsubsection{Training Technique for GPU Memory Limitations}
In our setup, the number of mini-batch per GPU denoted as $B$, is smaller than $N$ due to GPU memory size and fixed batch size limitations.
Therefore, we select $N$ AOT pairs on every GPU during our training iterations before sampling the training min-batch.
Subsequently, we sample each set of $B$ target images from the pre-selected $N$ pairs, as these pairs are stored in temporary data loaders.
 
\begin{figure*}[t]
\centering

\subfigure[]{\includegraphics[width=0.45\textwidth]{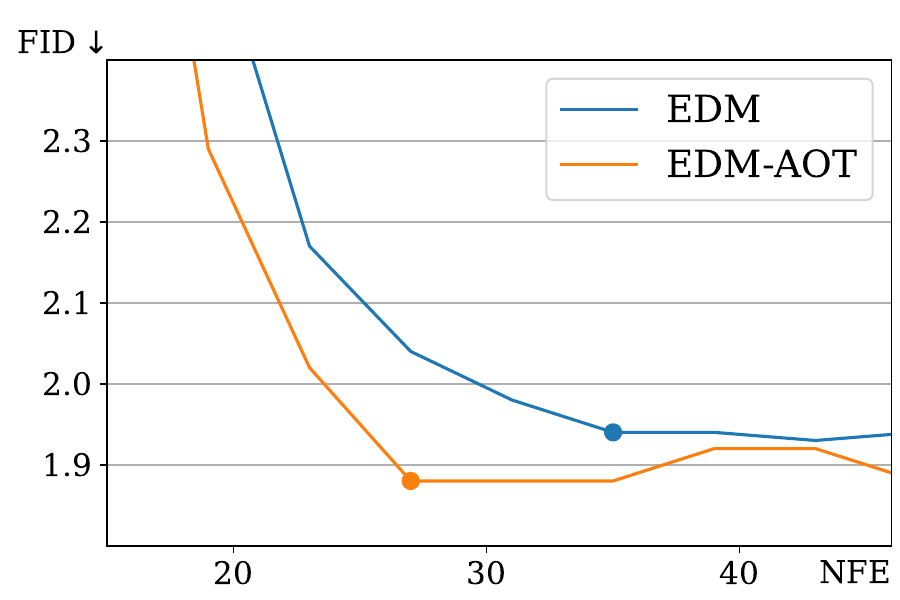}}
\hspace{2em}
\subfigure[]{\includegraphics[width=0.45\textwidth]{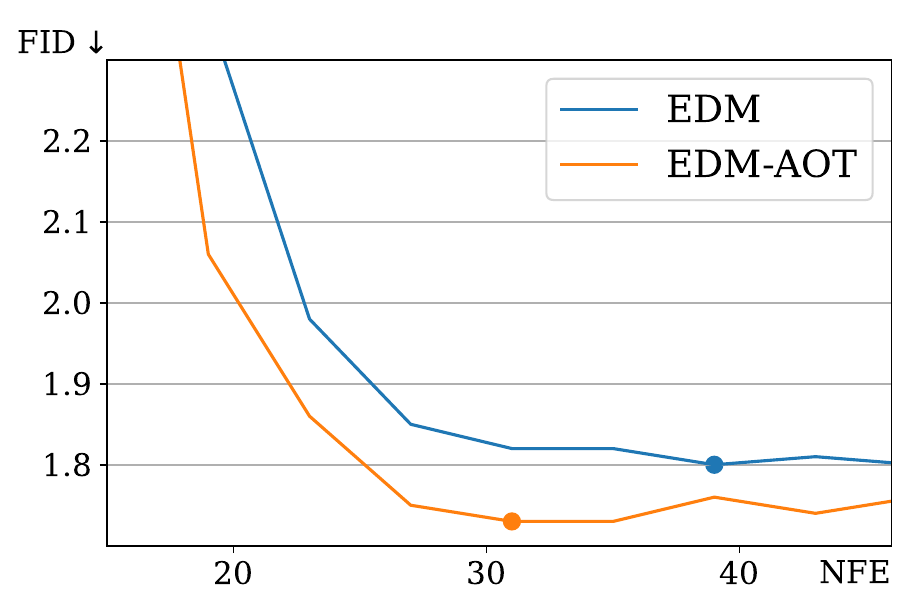}}
\vspace{-1em}
\caption{Variations in FID scores of CIFAR-10 images across different step counts. We sample images using four different $\rho$ values: 9, 27, 81, and 243, and select the optimal configuration. The dots on the graph represent points with minimal steps among those achieving the lowest FID score. (a) FID scores in unconditional generation. (b) FID scores in conditional generation.} 
\vspace{-1em}
\label{step_ablations}
\end{figure*}

\section{Experiments}

\subsection{Implementation Details}

 In our experiments using the CIFAR-10 dataset~\cite{krizhevsky2009learning} in an unconditional generation, we train models on 4 Nvidia Titan XP GPUs. The EDM framework serves as our baseline, with its hyper-parameters unchanged in our setup. Specifically, we use a batch per GPU, $B$, of 32, a batch size of 256, and set the number of AOT pairs, $N$, to 512.
 
 For the class-conditional generation on the CIFAR-10 dataset, we employ 4 Nvidia RTX 8000 GPUs. We adhere to the EDM's hyper-parameters, opting for a batch per GPU, $B$, of 32, a batch size of 128, and setting the number of AOT pairs, $N$, to 2048.

 We use the U-Net~\cite{ronneberger2015u} architecture used in the VE variant~\cite{song2021scorebased} of EDM. For evaluation, we use the FID score~\cite{heusel2017gans}; we sample 50,000 images once and conduct three separate measurements, reporting the lowest score obtained, following the EDM testing configurations.
 
\subsection{Comparison with Baselines}

In this experiment, we compare diffusion models trained using AOT and other diffusion-based generative models, which serve as our baselines.

Given that EDM models trained with our technique incorporate distinct ODE curvatures compared to prevailing EDM approaches, we explore optimal hyper-parameters, steps, and $\rho$ for the sampling process from Equation~\ref{equ-timesteps}.
Consequently, we provide results on the most favorable sampling performance achieved by selecting the hyper-parameters we meticulously identified.

Table~\ref{baseline},~\ref{baseline-cond}, and Figure~\ref{fig_main} present a comparative analysis of the results of other baseline methods. 
We denote the combination of the EDM model with our AOT technique as EDM-AOT.
The EDM-AOT model performs better than the baseline EDM approach under identical sampling configurations (18 steps, equivalent to 35 NFEs, and $\rho$ as 7).

Interestingly, we achieve enhanced performance even with fewer sampling steps, specifically 27 NFEs in unconditional generation and 29 NFEs in conditional generation, compared to alternative baseline methods.

\subsection{Analyzing Sampling Hyper-Parameters}
\label{rho_step}

In these experiments, we analyze the impact of the hyper-parameters, explicitly focusing on the NFEs and the $\rho$ values of Heun's sampler, as detailed in Equation~\ref{equ-timesteps}. 
We examine how these hyper-parameters influence the sampling performance concerning the EDM and EDM-AOT models.

\subsubsection{Analyzing the Impact of Sampling Stride}

We evaluate the sampling qualities in three settings: EDM with 35 NFEs, EDM-AOT with 35 NFEs, and EDM-AOT with 27 NFEs while varying the value of $\rho$. 
Figure~\ref{rho_ablation}-(a) and (b) illustrate the results of these experiments. 
In the unconditional generation, EDM achieves improved performance when employing slightly larger $\rho$ values ($\rho$ = 9 or 16, yielding FID scores of 1.94), compared to using the $\rho$ value utilized in EDM settings ($\rho$ = 7, an FID score of 1.97).

However, as $\rho$ increases further, the performance of EDM models deteriorates.
Conversely, diffusion models trained using the AOT technique exhibit results indicating that, as $\rho$ increases, the performances of EDM-AOT remain stable.
This tendency also holds in conditional generation.
Additionally, we conduct experiments with lower steps.
Figure~\ref{rho_ablation}-(c) presents the experiment outcomes.
In EDM, as the value of $\rho$ increases, the performance declines, while the model incorporating the AOT technique sustains its performance.

These results arise from the curvature of the ODE in diffusion models.
As the $\rho$ increases, the denoising strides in the sampling process become wider, particularly in the proximity of high noise levels.
In the case of EDM, the model estimates blurred images due to high information entropy, as depicted in Figure~\ref{1step_sample}-(b). 
In such instances, the utilization of high $\rho$ values magnifies the truncation error due to the utilization of a large stride for sampling in regions with high curvature.
This sampling approach results in a decrease in performance, as indicated by the experimental results.

On the other hand, the EDM-AOT model estimates more detailed images, as illustrated in Figure~\ref{1step_sample}-(c). Even with an increase in $\rho$, the truncation error remains minimal compared to EDM. Consequently, the performance at higher $\rho$ values remains consistent, and AOT-equipped models demonstrate remarkable performance by allocating multiple precise sampling steps within low noise levels.

\subsubsection{Analyzing the Impact of Sampling Steps}
In this experiment, we evaluate the performance of EDM and EDM-AOT across different numbers of steps.
However, as depicted in Figure~\ref{rho_ablation}, the optimal $\rho$ value, indicating the most effective sampling performance, varies depending on the models and sampling steps.
Therefore, we choose four $\rho$ values and sample images with these selected values, each executed with varying degrees of diversity steps. 
The chosen $\rho$ values are 9, 27, 81, and 243, all being powers of 3.

We present the optimal performance achieved by varying $\rho$ at each number of steps in Figure~\ref{step_ablations}. 
Diffusion models trained with the AOT technique maintain superior performance even with fewer steps. 
As the reduced curvature leads to minimal truncation errors in sampling, AOT-equipped diffusion models exhibit robustness even with lower sampling steps.

\subsection{Observing the Curvature of ODE Trajectory}

\begin{figure}
    \centering
    \subfigure[]{\includegraphics[width=0.985\columnwidth]{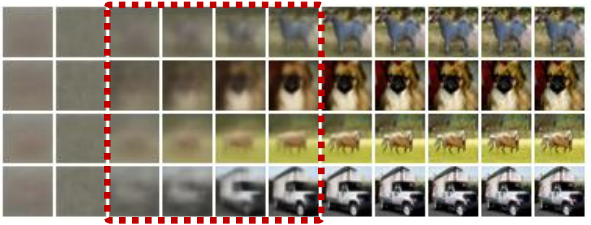}}
    
    \subfigure[]{\includegraphics[width=0.985\columnwidth]{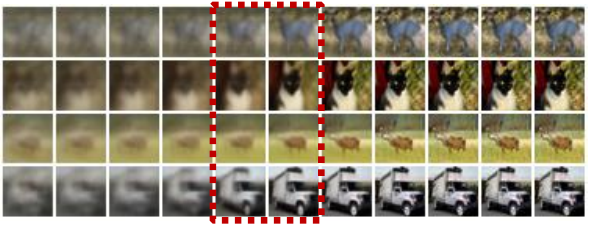}}
    \vspace{-1em}
    \caption{The images are generated through a single Euler step for each noisy image in the sampling process through unconditional generation, proceeding from left to right. The red boxes indicate the range where the generated images change. (a) Images generated using EDM. (b) Images generated using EDM-AOT.
    }
    \vspace{-1em}
    \label{ode_curve_comparison}
\end{figure}

Figure~\ref{ode_curve_comparison} illustrates the synthesized images obtained through a single Euler step in the sampling process using the EDM and EDM-AOT models.
We utilize the same random seed to sample Gaussian noise and generate images with the identical ODE integration stride, with 10 steps and $\rho$ as 7.

In Figure~\ref{ode_curve_comparison}-(a), the ODE trajectory of the EDM exhibits a broader range with higher curvature, leading to dynamic changes in the estimated images.
Meanwhile, EDM-AOT displays a narrower range of high curvature, as depicted in Figure~\ref{ode_curve_comparison}-(b). 
This characteristic implies that EDM-AOT has a lower truncation error, requiring fewer sampling iterations for image synthesis.

In addition, we observe that EDM and EDM-AOT produce similar images when subjected to the same noise. 
This phenomenon aligns with findings in several studies~\cite{song2023consistency,khrulkov2023understanding}. 
As emphasized in Song\textit{ et al.}~\shortcite{song2023consistency}, we can deduce that the AOT technique does not compromise the distinct sampling performance property of diffusion models. 
The independence in training between EDM and EDM-AOT reinforces this estimation.

\subsection{Discriminator Guidance with AOT}

Discriminator Guidance (DG) \cite{kim2022refining} in diffusion models enhances image quality by integrating a discriminator~\cite{goodfellow2020generative,karras2019style,karras2020analyzing} into the diffusion process. This discriminator guides the diffusion model's generation process using its gradients to synthesize images. This guidance helps boost the generation performance of diffusion models.

In the DG framework, the discriminator is typically trained using images generated by pre-trained diffusion models. In our experiments, we incorporate our EDM-AOT models within the DG framework. We train a discriminator using images generated by our EDM-AOT models. Significantly, we extend the application of the AOT technique to the discriminator's training process. By doing so, we use pairs of noise and images from the AOT procedure to synthesize noisy images for discriminator training. This integration of the AOT technique in the discriminator's training is a novel approach that aims to further enhance the performance of the diffusion model under discriminator guidance.

We gather a dataset of 50,000 images generated by our EDM-AOT models for these experiments.
The discriminator was trained with a batch size of 512 in unconditional generation, aligning with the number of pairs, $N$.
We determined 2048 pairs in training iterations for conditional generation, implementing the GPU-efficient method outlined in Section~\ref{4-2}.
This approach aims to leverage the benefits of AOT in refining discriminator performance, potentially leading to even higher-quality image generation.

Table~\ref{table2} shows the result.
The application of our EDM-AOT models demonstrates superior performance compared to the utilization of pre-trained EDM models.
Additionally, implementing AOT in the discriminator training yields robust performance gains in contrast to scenarios where these settings are not applied, resulting in notable FID scores of 1.68 in the unconditional generation and an FID score of 1.58 in the conditional generation with 29 NFEs.

\begin{table}[t]
\centering
\resizebox{.95\columnwidth}{!}{
\begin{tabular}{lccc}
\toprule
    Models & NFE $\downarrow$ & FID $\downarrow$  & FID $\downarrow$ \\
     & & (uncond)& (cond) \\ \midrule
    EDM w/o DG&35&1.97 & 1.79 \\
    EDM+DG&35&1.77 & 1.64\\ \midrule
    EDM-AOT w/o DG&\textbf{29}&1.88 & 1.73\\
    EDM-AOT + DG w/o AOT & \textbf{29} & 1.70 & 1.59 \\ 

    EDM-AOT + DG w/ AOT & \textbf{29} & \textbf{1.68} & \textbf{1.58}\\
    \bottomrule
\end{tabular}
}
\caption{The FID scores of CIFAR-10 images with discriminator guidance. We employ the AOT for the discriminator's training. ``DG w/ AOT" refers to the discriminator training with AOT. The combination of EDM-AOT and discriminator trained with AOT enhancement achieves state-of-the-art FID scores.}
\vspace{-1em}
\label{table2}
\end{table}

\section{Related Works}

In consideration of the fact that Song \textit{et al.}~\shortcite{song2021scorebased} model the diffusion sampling process using either SDE or ODE, substantial research has been dedicated to exploring the trade-offs associated with truncation errors.

Several studies have focused on model-centric and training-oriented strategies to mitigate these challenges.
\textbf{PFGM}~\cite{xu2022poisson} innovated by incorporating an electric field into the ODE, replacing the traditional diffusion process. This approach showed enhanced robustness in handling the ODE process. Further, \textbf{PFGM++}~\cite{xu2023pfgm++} established a linear relationship between PFGM and EDM, leading to considerable improvements in model performance.

Another line of research has explored using pre-trained models to enhance diffusion models, mainly through efficient samplers.
\textbf{PNDM}~\cite{liu2022pseudo} proposed replacing the Euler method with higher-order numerical integrators like the Runge-Kutta and linear multi-step methods~\cite{sauer2011numerical}. 
Similarly, \textbf{DEIS}~\cite{zhang2023fast} advocated for the application of Exponential Integrators~\cite{hochbruck2010exponential} as an efficient sampling method.

\section{Conclusion}

In this paper, we enhance the performance of diffusion models, facilitating the synthesis of high-quality images with reduced truncation errors by incorporating our proposed technique, AOT.
Additionally, our successful integration of AOT into the Discriminator Guidance (DG) framework showcases its versatility and potential for broader application.

However, we acknowledge certain limitations in our setup. 
Specifically, as per the experiment settings, we note a marginal increase in training costs (2\% to 15\%) compared to EDM.
Our approach requires algorithmic improvements to extend its applicability to challenging conditional generations, such as text guidance generation~\cite{saharia2022photorealistic,rombach2022high}.
These constraints offer valuable insights for our future endeavors and research investigations.
\clearpage

\bibliographystyle{named}
\bibliography{ijcai24}

\clearpage
\appendix

\begin{algorithm*}[t]
\caption{AOT algorithm for conditional generation}
\label{alg:assignment}

\begin{algorithmic}[1]
    \State \textbf{Input:} Images $\{y_i\}_{i=1}^{N}$, Noise $\{\epsilon_i\}_{i=1}^{N}$, Labels $\{l_i\}_{i=1}^{N}$

    \For{$i \in \{1,2, \dots, C\}$} \Comment{$C$: Number of classes in the dataset.}
        \State $S_i \leftarrow \{j \mid l_j == i\}$ 
        \Comment{$S_i$: Set of indexes where the label is the $i$-th class.}
    \EndFor
    
    \For{$i \in \{1,2, \dots, C\}$} 
    \Comment{Solve the assignment problem class-wise.}
        \State $\mathcal{Y} \leftarrow \{y_{j}\}_{j \in S_i}$ 
       \Comment{$\mathcal{Y}$: Temporary subset of images sharing the same label.}
        \State $\mathcal{E} \leftarrow \{\epsilon_{j}\}_{j \in S_i}$ 
        \Comment{$\mathcal{E}$: Temporary subset of noise sharing the same label.}
        \State \textbf{Find} $\pi \leftarrow  \arg\min_\pi \sum_{i=1}^N c (\mathcal{E}^\pi_i, \mathcal{Y})_i$
        \Comment{Solve the class-wise assignment problem using the Hungarian algorithm.}
        \For{$k \in S_i$}
        \Comment{Maintain the index of images to preserve randomness.}
            \State $\epsilon'_k \leftarrow \mathcal{E}^\pi_k$ 
             \Comment{Update noise into the selected noise $\epsilon'$.}
        \EndFor
    \EndFor
    \State \textbf{Output:} Selected noise $\epsilon'$
\end{algorithmic}
\end{algorithm*}

    \begin{table*}[]
    \centering
    \begin{tabular}{ccccccccccc}
    \toprule
        $\rho$ &  3&5&9&16&27&47& 81& 140& 243& 421\\ \midrule
       EDM, 18 steps, uncond  &6.16 &  1.99&    1.94  &  1.94  &  1.99 &  2.00 &  2.01  &  2.02  &  1.99  &  1.98 \\
       EDM-AOT, 18 steps, uncond& 6.97  &  2.18  & 1.93 &   2.06   & 2.11   & 1.90 &1.93  &  1.89 &   1.89  &  1.89 \\
       EDM-AOT, 14 steps, uncond &18.77   & 2.37  &  2.32   & 1.95  &  1.93  &  1.94    &1.88    &1.88 &   1.89    &1.88 \\ \midrule
       EDM, 18 steps, cond&6.15 &  1.85&    1.82  &  1.8  &  1.82 &  1.88 &  1.88  &  1.89  &  1.89  &  1.88 \\
       EDM-AOT, 18 steps, cond&6.98  &  1.88  & 1.77 &   1.78   & 1.76   &1.74 &1.73 &1.73 &  1.74 &   1.74\\
       EDM-AOT, 14 steps, cond& 17.68   & 2.18  &  1.85   & 1.77  &  1.76  &  1.76    &1.76   &1.75 &   1.75    &1.75\\ \midrule
       EDM, 8 steps, uncond&89.13 &  9.99&    4.05  &  4.08  &  4.83 &  5.43 &  5.98   &  6.16& 6.17  &  6.28\\
       EDM-AOT, 8 steps, uncond&90.58  &  10.92  & 3.78 &   3.24  &3.06  & 3.11 &3.06 &  3.14 &   3.15  & 3.15\\  \bottomrule
    \end{tabular}
    \caption{Detailed results corresponding to Figure~\ref{rho_ablation}.}
    \label{fig4_data}
\end{table*}

\begin{table*}[]
    \centering
    \begin{tabular}{cccccccccccc}
    \toprule
      NFE   & 7&11&15&19&23&27&31&35&39& 43&47 \\ \midrule
      EDM, uncond   & 68.98&11.86& 4.05& 2.58&2.17&2.04&1.98&1.94& 1.94&1.93&1.94 \\
      EDM-aot, uncond &58.55&7.51&3.06&2.29&2.02&1.88&1.88&1.88& 1.92&1.92&1.88 \\ \midrule
      EDM, cond & 63.32&11.06& 3.71& 2.36&1.98&1.85&1.82&1.82& 1.80
&1.81&1.80\\
EDM-AOT, cond&61.36&7.65&2.91&2.06&1.86&1.75&1.73&1.73
& 1.76&1.74&1.76 \\ \bottomrule
    \end{tabular}
    \caption{Detailed results corresponding to Figure~\ref{step_ablations}.}
    \label{fig5_data}
\end{table*}

\begin{figure*}
    \centering
    \subfigure[EDM]{\includegraphics[width = 0.7\textwidth]{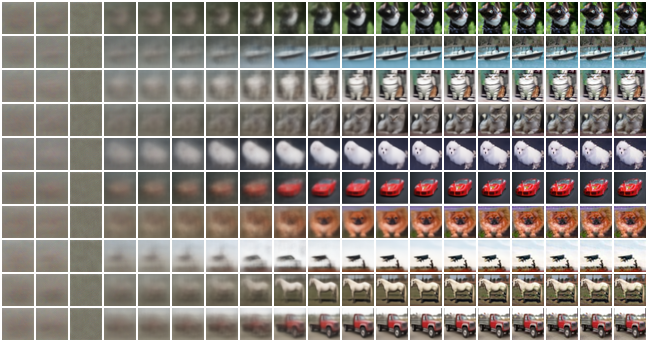}}    
    \subfigure[EDM-AOT]{\includegraphics[width = 0.7\textwidth]{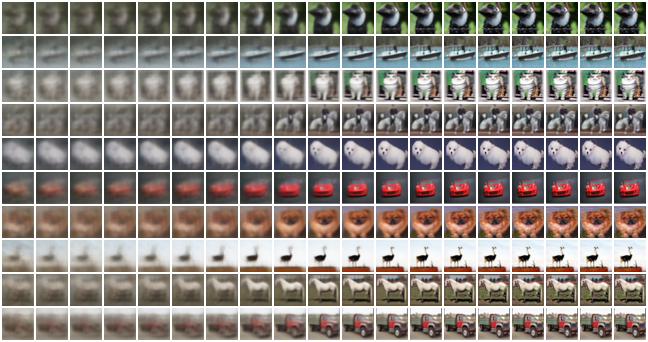}}    

    \caption{The images are generated using the unconditional models with a single Euler step at each noisy image during the sampling process, proceeding from left to right. We use 18 steps (35 NFEs) and $\rho$ as 7.}
    \label{fifif}
\end{figure*}

\begin{figure*}
    \centering

    \subfigure[EDM]{\includegraphics[width = 0.7\textwidth]{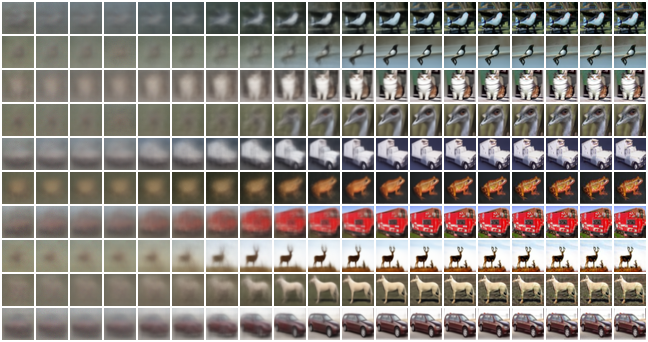}}    
    \subfigure[EDM-AOT]{\includegraphics[width = 0.7\textwidth]{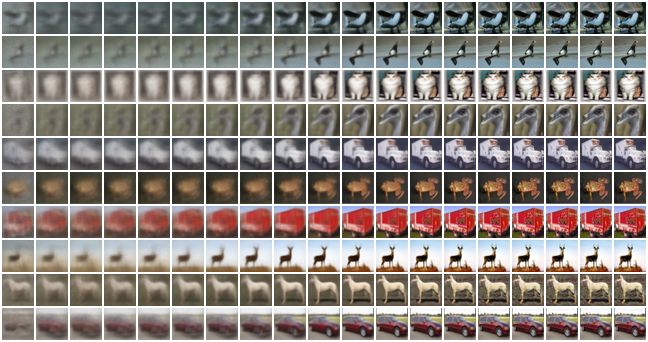}}
    \caption{The images are generated using the conditional models with a single Euler step at each noisy image during the sampling process, proceeding from left to right. We use 18 steps (35 NFEs) and $\rho$ as 7.}
    \label{fig:enter-label}
\end{figure*}

\begin{figure*}
    \centering
    \subfigure[EDM, deer]{\includegraphics[width = 0.4\textwidth]{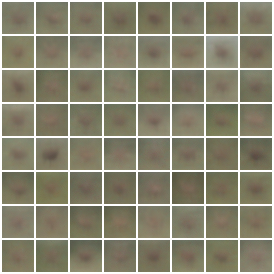}}
    \subfigure[EDM-AOT, deer]{\includegraphics[width = 0.4\textwidth]{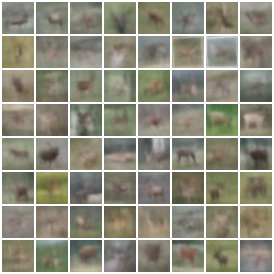}}
        \subfigure[EDM, truck]{\includegraphics[width = 0.4\textwidth]{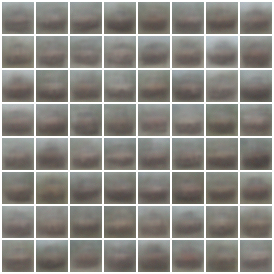}}
    \subfigure[EDM-AOT, truck]{\includegraphics[width = 0.4\textwidth]{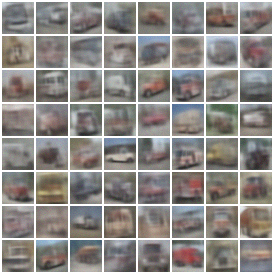}}
    \caption{Denoised CIFAR-10 images $\mathbf{x}_0$ in conditional generation using single-step Euler method. }
    \label{fig:enter-label}
\end{figure*}

\begin{figure*}
    \centering
    \subfigure[18 steps, $\rho$ as 7, FID score of 1.95]{\includegraphics[width=0.3\textwidth]{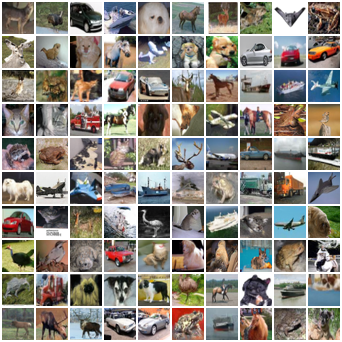}}
    \subfigure[14 steps, $\rho$ as 90, FID score of 1.88]{\includegraphics[width=0.3\textwidth]{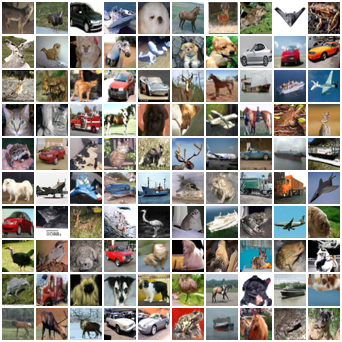}}
    \subfigure[15 steps, $\rho$ as 60, DG, FID score of 1.68]{\includegraphics[width=0.3\textwidth]{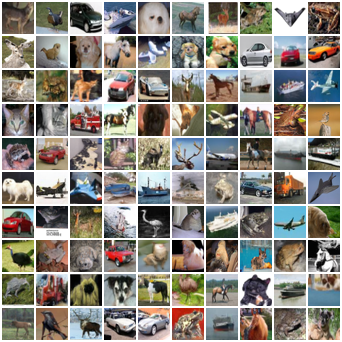}}
        \caption{Synthesized images using unconditional EDM-AOT with different settings.}
\end{figure*}

\begin{figure*}
    \centering
    \subfigure[18 steps, $\rho$ as 7, FID score of 1.79]{\includegraphics[width=0.3\textwidth]{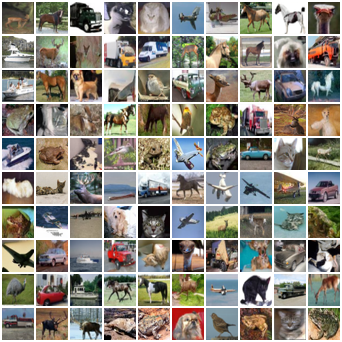}}
    \subfigure[15 steps, $\rho$ as 72, FID score of 1.73]{\includegraphics[width=0.3\textwidth]{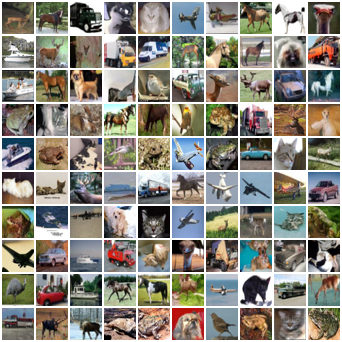}}
        \subfigure[15 steps, $\rho$ as 72, DG, FID score of 1.58]{\includegraphics[width=0.3\textwidth]{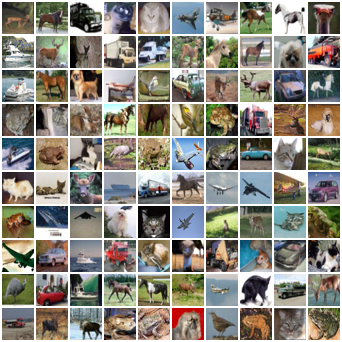}}
        \caption{Synthesized images using conditional EDM-AOT with different settings.}
\end{figure*}

\end{document}